\documentclass[lettersize,journal]{IEEEtran}

\usepackage{amsmath,amsfonts}
\usepackage{algpseudocode}
\usepackage{algorithmicx,algorithm}
\usepackage{hyperref}
\usepackage{array}
\usepackage[caption=false,font=normalsize,labelfont=sf,textfont=sf]{subfig}
\usepackage{textcomp}
\usepackage{stfloats}
\usepackage{url}
\usepackage{verbatim}
\usepackage{graphicx}
\usepackage{cite}

\usepackage[square,sort&compress,numbers]{natbib}
\usepackage{amsmath}
\usepackage{amsfonts} 
\usepackage{amssymb} 
\usepackage{multirow}
\usepackage{threeparttable}
\usepackage{booktabs}

\usepackage{bbm, dsfont}
\usepackage{bbding}
\usepackage{ifsym}

\graphicspath{{figs/}}
\graphicspath{{bio/}}

\graphicspath{{figs/}}

\begin{document}
\title{RSMamba: Remote Sensing Image Classification with State Space Model}

\author{Keyan~Chen$^{1}$,~Bowen~Chen$^{1}$,~Chenyang~Liu$^{1}$,~Wenyuan~Li$^{2}$,~Zhengxia~Zou$^{1}$,~Zhenwei~Shi$^{1, \star}$
\\
\vspace{6pt}
Beihang University$^1$, The University of Hong Kong$^2$
}

\maketitle

\begin{abstract}

Remote sensing image classification forms the foundation of various understanding tasks, serving a crucial function in remote sensing image interpretation. 
The recent advancements of Convolutional Neural Networks (CNNs) and Transformers have markedly enhanced classification accuracy.
Nonetheless, remote sensing scene classification remains a significant challenge, especially given the complexity and diversity of remote sensing scenarios and the variability of spatiotemporal resolutions. 
The capacity for whole-image understanding can provide more precise semantic cues for scene discrimination.
In this paper, we introduce RSMamba, a novel architecture for remote sensing image classification. RSMamba is based on the State Space Model (SSM) and incorporates an efficient, hardware-aware design known as the Mamba. It integrates the advantages of both a global receptive field and linear modeling complexity. To overcome the limitation of the vanilla Mamba, which can only model causal sequences and is not adaptable to two-dimensional image data, we propose a dynamic multi-path activation mechanism to augment Mamba's capacity to model non-causal data. Notably, RSMamba maintains the inherent modeling mechanism of the vanilla Mamba, yet exhibits superior performance across multiple remote sensing image classification datasets. This indicates that RSMamba holds significant potential to function as the backbone of future visual foundation models. The code will be available at \url{https://github.com/KyanChen/RSMamba}.

\end{abstract}


\IEEEpeerreviewmaketitle

\section{Introduction}

\IEEEPARstart{T}he advancement of remote sensing technology has significantly heightened interest in high-resolution earth observation. Remote sensing image classification, serving as the bedrock of remote sensing image intelligent interpretation, is a crucial element for subsequent downstream tasks. It plays a pivotal role in applications such as land mapping, land use, and urban planning. Nonetheless, the complexity and diversity of remote sensing scenarios, coupled with the variable spatio-temporal resolution, present substantial challenges to automated remote sensing image classification \cite{xia2017aid, yang2010bag, cheng2017remote,chen2022resolution}.

Researchers have been diligently working towards alleviating these challenges and enhancing the models' applicability across diverse application scenarios. Early methodologies predominantly focused on feature construction, extraction, and selection, investigating feature engineering machine learning methods represented by SIFT, LBP, color histograms, GIST, BoVW \cite{li2018deep}, \textit{etc}. 
In recent years, the advent of deep learning has revolutionized the conventional paradigm that heavily relied on specialized human prior knowledge. Deep learning possesses the capability to autonomously mine effective features from data and output classification probabilities in an end-to-end manner. In terms of network architecture, it can primarily be categorized into CNNs and attention networks. The former abstracts image features layer by layer through two-dimensional convolution operations, as demonstrated by ResNet \cite{he2016deep}. The latter captures long-distance dependencies between local areas of the entire image through the attention mechanism, thereby achieving a more robust semantic response, represented by ViT \cite{dosovitskiy2020image}, SwinTransformer \cite{liu2021swin}, \textit{etc}.
Substantial progress has also been made in remote sensing image classification. For instance, ET-GSNet \cite{xu2022vision} distills the rich semantic prior of ViT into ResNet18, fully capitalizing on the strengths of both. P2Net \cite{chen2022contrastive} introduces an asynchronous contrastive learning method to address the issue of small inter-class differences in fine-grained classification.

To a certain extent, the classification accuracy heavily depends on the model's ability to effectively handle the impact of complex and diverse remote sensing scenarios and variable spatio-temporal resolution. Transformer \cite{vaswani2017attention}, based on the attention mechanism and capable of obtaining responses from valuable areas across the entire image, presents an optimal solution to these challenges. However, its attention calculation, characterized by square complexity, poses significant challenges in terms of modeling efficiency and memory usage as the input sequence length increases or the network deepens. The State Space Model (SSM) \cite{gu2021efficiently} can establish long-distance dependency relationships through state transitions and execute these transitions via convolutional calculations, thereby achieving near-linear complexity. Mamba \cite{gu2023mamba} proves highly efficient for both training and inference by incorporating time-varying parameters into the plain SSM and conducting hardware optimization. Vim \cite{zhu2024vision} and VMamba \cite{liu2024vmamba} have successfully introduced Mamba into the two-dimensional visual domain, achieving a commendable balance of performance and efficiency across multiple tasks.

In this paper, we introduce RSMamba, an efficient state space model for remote sensing image classification. 
Owing to its robust capability in modeling global relationships within an entire image, RSMamba can also exhibit potential versatility across a broad spectrum of other tasks. 
RSMamba is based on the previous Mamba \cite{gu2023mamba}, but has introduced a dynamic multi-path activation mechanism to alleviate the limitations of the plain Mamba, which can only model in a single direction and is position-agnostic. Significantly, RSMamba is designed to preserve the inherent modeling mechanism of the original Mamba block, while introducing non-causal and position-positive improvements external to the block. Specifically, the remote sensing image is partitioned into overlapping patch tokens, to which position encoding is added to form a sequence. We construct three path copies, namely forward, reverse, and random. These sequences are modeled to incorporate global relationships through the Mamba block using shared parameters, and subsequently activated through linear mapping across different paths. Given the efficiency of the Mamba block, large-scale pre-training of RSMamba can be achieved cost-effectively.

\vspace{6pt}
The primary contributions of this paper can be summarized as follows:

i) We propose RSMamba, an efficient global feature modeling methodology for remote sensing images based on the State Space Model (SSM). This method offers substantial advantages in terms of representational capacity and efficiency and is expected to serve as a feasible solution for handling large-scale remote sensing image interpretation.

ii) Specifically, we incorporate a position-sensitive dynamic multi-path activation mechanism to address the limitation of the original Mamba, which was restricted to modeling causal sequences and was insensitive to the spatial position.

iii) We conducted comprehensive experiments on three distinct remote sensing image classification datasets. The results indicate that RSMamba holds significant advantages over classification methods based on CNNs and Transformers.

\begin{figure*}[t]
\centering
% \resizebox{宽度}{高度}{对象}
\resizebox{0.99\linewidth}{!}{
\includegraphics[width=\linewidth]{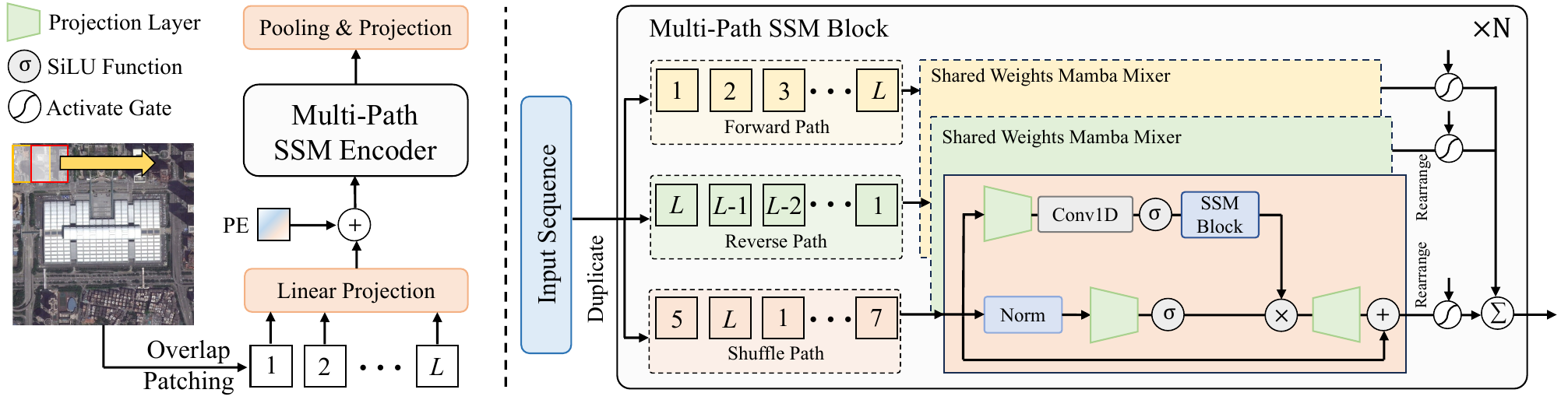}
}
\vspace{-6pt}
\caption{An overview of the proposed RSMamba.
\label{fig:model}}
\vspace{-10pt}
\end{figure*}

\section{Methodology}

Leveraging the inherent characteristics of the SSM model, RSMamba is proficient in effectively capturing the global dependencies within remote sensing images, thereby yielding a wealth of semantic category information. This section will begin with an introduction to the preliminaries of SSM, followed by an overview of RSMamba. Subsequently, we will explore the dynamic multi-path activation block in depth. Finally, we will elaborate on the network structure for three distinct versions of RSMamba.

\subsection{Preliminaries}

The State Space Model (SSM) is a concept derived from modern control theory's linear time-invariant system which maps the continuous stimulation $x \in \mathbb{R}^N$ to response $y \in \mathbb{R}^N$. This process can be formulated through the subsequent linear ordinary differential equation (ODE),
\begin{equation}
    \begin{aligned}
        h^\prime(t) &= \textbf{A} h(t) + \textbf{B} x(t) \\
        y(t) &= \textbf{C} h(t)
    \end{aligned}
    \label{eq:continous_sys}
\end{equation}
where $y \in \mathbb{R}^N$ is derived from the input signal $x \in \mathbb{R}^N$ and the hidden state $h \in \mathbb{R}^N$. $\textbf{A}  \in \mathbb{R}^{N \times N}$ denotes the state transition matrix. $\textbf{B} \in \mathbb{R}^{N}$ and $\textbf{C} \in \mathbb{R}^{ N}$ are the projection matrices. 
To realize the continuous system depicted in Eq. \ref{eq:continous_sys} in a discretized form and integrate it into deep learning methods. $\textbf{A}$ and  $\textbf{B}$ are discretized using a zero-order hold (ZOH) with a time scale parameter $\Delta$. The process is shown as follows,
\begin{equation}
    \begin{aligned}
        \bar{\textbf{A}} &= \text{exp}(\Delta \textbf{A}) \\
        \bar{\textbf{B}} &= {(\Delta \textbf{A})}^{-1} (\text{exp}(\Delta \textbf{A}) - \textbf{I}) \cdot \Delta \textbf{B} \\
    \end{aligned}
\end{equation}

After discretization, Eq. \ref{eq:continous_sys} can be rewritten as,
\begin{equation}
    \begin{aligned}
        h_k &= \bar{\textbf{A}} h_{k-1} + \bar{\textbf{B}} x_k \\
        y_k &= \bar{\textbf{C}} h_k
    \end{aligned}
\end{equation}
where $\bar{\textbf{C}}$ represents $\textbf{C}$. At last, the output can be calculated in a convolution representation, as follows,
\begin{equation}
    \begin{aligned}
        \bar{\textbf{K}} &= (\bar{\textbf{C}} \bar{\textbf{B}}, \bar{\textbf{C}} \bar{\textbf{A}} \bar{\textbf{B}}, \cdots, \bar{\textbf{C}} \bar{\textbf{A}}^{L-1} \bar{\textbf{B}}) \\
        \textbf{y} &= \textbf{x} \ast \bar{\textbf{K}} 
    \end{aligned}
\end{equation}
where $L$ is the length of the input sequence, and $\bar{\textbf{K}} \in \mathbb{R}^L$ denotes the structured convolutional kernel.

\subsection{RSMamba}

RSMamba transforms 2-D images into 1-D sequences and captures long-distance dependencies using the Multi-Path SSM Encoder, as depicted in Fig. \ref{fig:model}. Given an image $\mathcal{I} \in \mathbb{R}^{H \times W \times 3}$, we employ a 2-D convolution with a kernel of $k$ and a stride of $s$ to map local patches into pixel-wise feature embeddings. Subsequently, the feature map is flattened into a 1-D sequence. To preserve the relative spatial position relationship within the image, we incorporate position encoding $P$. The entire process is as follows,
\begin{equation}
    \begin{aligned}
    T &= \Phi_{\text{Flatten}}(\Phi_{\text{Conv2D}}(\mathcal{I}, k, s)) \\
    T &= T + P
    \end{aligned}
\end{equation}
where $\Phi_{\text{Conv2D}}$ represents the 2-D convolution, while $\Phi_{\text{Flatten}}$ signifies flattening operation.  $T \in \mathbb{R}^{L \times d}$ and $P \in \mathbb{R}^{L \times d}$ correspond to the input 1-D sequence and positional encoding, respectively.

In RSMamba, we have not utilized the [CLS] token to aggregate the global representation, as is done in ViT. Instead, the sequence is fed into multiple dynamic multi-path activation Mamba blocks for long-distance dependency modeling. Subsequently, the dense features necessary for category prediction are derived through a mean pooling operation applied to the sequence. This procedure can be iteratively delineated as follows,
\begin{equation}
    \begin{aligned}
    T^i &= \Phi_{\text{mp-ssm}}^{i} (T^{i-1}) + T^{i-1}\\
    \hat{s} &= \Phi_{\text{proj}}(\Phi_{\text{LN}}(\Phi_{\text{mean}}(T^N)))\\
    \end{aligned}
\end{equation}
where $i$ signifies the $i$th layer, while $T^i$ represents the output sequence of the $i$th-layer, with $T^0 = T  \in \mathbb{R}^{L \times d}$. $\Phi_{\text{mp-ssm}}$ denotes the dynamic multi-path activation Mamba block, with a total number of $N$. $\Phi_{\text{mean}}$ symbolizes mean pooling operation with the sequence dimension and $\Phi_{\text{LN}}$ is layer normalization. $\Phi_{\text{proj}}$ is used to project the latent dimension $d$ to the number of classes.

\subsection{Dynamic Multi-path Activation}

The vanilla Mamba is employed for the causal modeling of 1-D sequences. It encounters difficulties in modeling spatial positional relationships and unidirectional paths, thereby limiting the applicability to visual data representation. To augment its capacity for 2-D data, we introduce a dynamic multi-path activation mechanism. Importantly, this mechanism, to preserve the structure of the vanilla Mamba block, exclusively operates on the block's input and output. Specifically, we duplicate three copies of the input sequence to establish three different paths, namely the forward path, reverse path, and random shuffle path, and leverage a plain Mamba mixer with shared parameters to model the dependency relationships among tokens within these three sequences, respectively. Subsequently, we revert all tokens in the sequences to the correct order and employ a linear layer to condense sequence information, thereby establishing the gate of the three paths. This gate is then used to activate the representation of the three different information flows as shown in Fig. \ref{fig:model}. The process of the $i$th block is delineated as follows,
\begin{equation}
    \begin{aligned}
    T_k^i &= \Phi_{\text{pather}}^k(T^i) \\
    \hat{T}_k^i &= \Phi_{\text{mixer}}^\theta(E_k^i) \\
    \hat{T}_k^i &= \Phi_{\text{revert-pather}}^k(\hat{E}_k^i) \\
    g &= \Phi_{\text{softmax}}(\Phi_{\text{gate-proj}}(\Phi_{\text{mean}}(\Phi_{\text{cat}}(\{\hat{E}_k^i \})))) \\
    T^{i+1} &= \sum\nolimits_{k=0}^2 g_k \cdot \hat{T}_k^i \\
    \end{aligned}
\end{equation}
where $T^i$ represents the input sequence for the $i$th layer. $\Phi_{\text{pather}}^k, k \in \{0,1,2 \}$ denotes the $k$th sequence path, including the forward path, reverse path, and random shuffle path. $\Phi_{\text{mixer}}^\theta$ is the vanilla Mamba mixer with parameter $\theta$. $\Phi_{\text{revert-pather}}^k$ denotes the operation to revert all tokens to the forward order. $\Phi_{\text{cat}}$ signifies sequence concatenation with the feature dimension. $\Phi_{\text{mean}}$ denotes mean pooling along the sequence length dimension. $\Phi_{\text{gate-proj}}$ linearly projects the $3d$ dimension to 3 for sequence information activation. $\Phi_{\text{softmax}}$ denotes Softmax operation. $\sum$ gathers features from the three different information flows.

\subsection{Model Architecture}

The Mamba mixer $\Phi_{\text{mixer}}^\theta$ represents the standard mixer block within the Mamba \cite{gu2023mamba} framework. Drawing upon the principles of ViT, we have developed three distinct versions of RSMamba characterized by different parameter sizes: base, large, and huge. The specific hyperparameters for each version are detailed in Tab. \ref{tab:versions}. Details about the hyperparameter meaning can be found in \cite{gu2023mamba}.

\begin{table}[!tbhp] 
\centering
\caption{The hyperparameter settings for different RSMamba versions. N: Number of blocks, HS: Hidden Size, IS: Intermediate Size, TSR: Time Step Rank, SSMSS: SSM State Size.
}
\label{tab:versions}
\resizebox{0.8\linewidth}{!}{
\begin{tabular}{c| *5{c}}
\toprule
Version & N & HS & IS  & TSR & SSMSS
\\
\midrule
Base & 24 & 192 & 384 & 12 & 16
\\
Large & 36 & 256 & 512 & 16 & 16
\\
Huge & 48 & 320 & 640 & 20 & 16 
\\
\bottomrule
\end{tabular}
}
\end{table}

\begin{table*}[!tbhp] 
\centering
\caption{Comparisons with other methods across different test sets.
}
\label{tab:sota}

\resizebox{0.9\linewidth}{!}{
\begin{tabular}{*2{c}| *3{c}| *3{c}|*3{c} }
\toprule
\multirow{2}{*}{Method} & Params & \multicolumn{3}{c|}{UC Merced} & \multicolumn{3}{c|}{AID} & \multicolumn{3}{c}{RESISC45} 
\\
& (M) & P & R & \multicolumn{1}{c|}{F1} & P & R & \multicolumn{1}{c|}{F1} & P & R & F1
\\
\midrule
ResNet-18 \cite{he2016deep} & 11.2 
& 87.98 & 87.46 & 87.40 
& 88.70 & 88.17 & 88.30 
& 88.73 & 88.44 & 88.45
\\
ResNet-50 \cite{he2016deep} & 23.6
& 91.99 & 91.74 & 91.65 
& 89.44 & 88.66 & 88.87 
& 92.67 & 92.47 & 92.47
\\
ResNet-101 \cite{he2016deep} & 42.6
& 92.40 & 92.22 & 92.12 
& 91.03 & 90.63 & 90.81 
& 92.75 & 92.57 & 92.56
\\
\midrule
DeiT-T \cite{touvron2021training} & 5.5 
& 86.92 & 86.66 & 86.53 
& 85.23 & 84.52 & 84.52
& 87.66 & 86.78 & 86.79
\\
DeiT-S \cite{touvron2021training} & 21.7 
& 88.95 & 88.41 & 88.41 
& 85.88 & 85.19 & 85.34
& 88.21 & 87.47 & 87.43
\\
DeiT-B \cite{touvron2021training} & 85.8 
& 89.14 & 88.73 & 88.70 
& 87.32 & 86.07 & 86.07
& 89.04 & 88.62 & 88.65
\\
ViT-B \cite{dosovitskiy2020image} & 88.3
& 91.09 & 90.79 & 90.77 
& 89.39 & 88.65 & 88.86
& 88.84 & 88.65 & 88.62
\\
ViT-L \cite{dosovitskiy2020image} & 303.0 
& 91.98 & 91.32 & 91.26 
& 90.19 & 88.86 & 89.17
& 91.22 & 91.08 & 91.04
\\
Swin-T \cite{liu2021swin} & 27.5 
& 90.87 & 90.63 & 90.40 
& 86.49 & 85.66 & 85.77 
& 90.15 & 90.06 & 90.06
\\
Swin-S \cite{liu2021swin} &48.9 
& 91.08 & 90.95 & 90.82 
& 87.50 & 86.80 & 86.89 
& 92.05 & 91.88 & 91.84
\\
Swin-B \cite{liu2021swin} & 86.8 
& 91.85 & 91.74 & 91.62 
& 89.84 & 89.01 & 89.07 
& 93.63 & 91.58 & 93.56
\\
\midrule
Vim-Ti$^\dag$ \cite{zhu2024vision} & 7.0 
& 89.06 & 88.73 & 88.68
& 87.76 & 86.98 & 87.13
& 89.24 & 89.02 & 88.97
\\
VMamba-T \cite{liu2024vmamba} & 30.0 
& 93.14 & 92.85 & 92.81
& 91.59 & 90.94 & 91.10
& 93.97 & 93.96 & 93.94
\\
\midrule
RSMamba-B (Ours) & 6.4 
& 94.14 & 93.97 & 93.88 
& 92.02 & 91.53 & 91.66
& 94.87 & 94.87 & 94.84
\\
RSMamba-L (Ours) & 16.2 
& 95.03 & 94.76 & 94.74 
& 92.31 & 91.75 & 91.90
& 95.03 & 95.05 & 95.02
\\
RSMamba-H (Ours) & 33.1 
& \textbf{95.47} & \textbf{95.23} & \textbf{95.25} 
& \textbf{92.97} & \textbf{92.51} & \textbf{92.63} 
& \textbf{95.22} & \textbf{95.19} & \textbf{95.18}
\\
\bottomrule
\end{tabular}
}
\end{table*}

\section{Experimental Results and Analyses}
\subsection{Dataset Description}

To evaluate the efficacy of the proposed method, we undertook extensive experiments on three distinct remote datasets: UC Merced Land-Use Dataset (UC Merced) \cite{yang2010bag}, AID \cite{xia2017aid}, and NWPU-RESISC45 Dataset (RESISC45) \cite{cheng2017remote}. Each encompasses a unique assortment of categories and image quantities.

\textbf{UC Merced} \cite{yang2010bag}: The UC Merced is composed of 21 distinct scene categories, with each category containing 100 aerial images of $256 \times 256$ pixel resolution. The images possess a spatial resolution of 0.3m, culminating in a total of 2100 images. We randomly extracted 70 images from each category for training.

\textbf{AID} \cite{xia2017aid}: The AID incorporates 30 categories and an aggregate of 10,000 images sourced from Google Earth. The sample quantity varies across different scene types, ranging from 220 to 420. Each aerial image measures $600 \times 600$ pixels, with spatial resolutions spanning from 8m to 0.5m, thereby encapsulating a multitude of resolution scenarios. We designated 50\% of the images from each category as training data.

\textbf{RESISC45} \cite{cheng2017remote}: The RESISC45 comprises 31,500 remote sensing images obtained from Google Earth, segregated into 45 scene categories. Each category contains 700 RGB images with $256 \times 256$ pixel resolution. The spatial resolution fluctuates between approximately 30m to 0.2m per pixel. We allocated 70\% of the images from each category for training purposes.

\subsection{Implementation Details}

In our paper, we employ a fixed input image size of $224 \times 224$ and implement data augmentation techniques including random cropping, flipping, photometric distortion, mixup, cutMix, \textit{etc}. Images are processed into sequential data through a two-dimensional convolution with a kernel size of 16 ($k=16$) and a stride of 8 ($s=8$). Position encodings are represented by randomly initialized learnable parameters. For supervised training, we employ the cross-entropy loss function and utilize the AdamW optimizer with an initial learning rate of $5e-4$ and a weight decay of 0.05. The learning rate is decayed using a cosine annealing scheduler with a linear warmup. The batch size for training is set at 1024, and the training process spans a total of 500 epochs. We employ Precision (P), Recall (R), and F1-score (F1) as performance metrics.

\subsection{Comparison with the State-of-the-Art}

We compare our proposed RSMamba with other prevalent deep learning methods for image classification, including the ResNet \cite{he2016deep} series underpinned by CNN architecture, and the DeiT \cite{touvron2021training}, ViT \cite{dosovitskiy2020image}, and Swin Transformer \cite{liu2021swin} series, all of which are grounded in Transformer architecture. The comparative classification performance of these methods across the UC Merced, AID, and RESISC45 datasets is presented in Tab. \ref{tab:sota}.
The experimental results reveal that: i) RSMamba exhibits robust performance across datasets of varying sizes, with its efficacy being minimally impacted by the volume of training data. This could be attributed to its relatively fewer parameters, negating the need for extensive data for inductive bias. ii) An increase in the depth and width of RSMamba contributes to a performance enhancement across the three datasets. However, the rate of improvement is less pronounced compared to the ResNet and Transformer series. This could be because the base version of RSMamba has already achieved a high degree of accuracy relative to other methods, suggesting that the base version could be a viable starting point for other application tasks. iii) Our experiments also indicate that while CNN architectures converge readily, the superior performance of Transformer architectures hinges on the induction and bias of general features across large-scale training data. In contrast, RSMamba's performance does not rely on extensive data accumulation, but a longer training duration can further lead to substantial performance gains.

\subsection{Ablation Study}

To verify the effectiveness of each component, ablation experiments were conducted on the AID dataset. Unless explicitly stated, the base version of the model was utilized, with no modifications made to the associated hyperparameters.

\subsubsection{Effect of Class Tokens}

To obtain dense semantic features for classification, we leveraged mean pooling in RSMamba to amalgamate global information, as opposed to using class tokens akin to ViT \cite{dosovitskiy2020image}. Tab. \ref{tab:ablation-clstoken} delineates the effect of incorporating class tokens at varying positions and mean pooling on the classification performance. The experimental findings indicate that the insertion of class tokens at the head, tail, or both does not yield superior performance. However, insertion in the middle of the sequence can result in a substantial enhancement in performance. Moreover, mean pooling on the sequence can exhibit optimal performance. These observations suggest that the direction of information flow in Mamba significantly influences performance. Concurrently, it was observed during the experiment that mean pooling can expedite the network's convergence.

\begin{table}[!tbhp] 
\centering
\caption{Effect of class tokens and mean pooling on performance.
}
\label{tab:ablation-clstoken}
\vspace{-6pt}
\resizebox{\linewidth}{!}{
\begin{tabular}{cccc | *3{c}}
\toprule
Head & Tail & Middle & Mean Pooling  & P & R & F1
\\
\midrule
\checkmark& & &  & 87.71 & 86.71 & 86.92
\\
& \checkmark& &  & 88.68 & 87.58 & 87.74
\\
\checkmark &\checkmark & & & 87.92 & 86.35 & 86.79
\\
& &\checkmark &  & 91.63 & 91.19 & 91.24
\\
& & & \checkmark & 92.02 & 91.53 & 91.66
\\
\bottomrule
\end{tabular}
}
\end{table}

\subsubsection{Effect of Multiple Scanning Paths}

The vanilla Mamba, derived from modeling causal sequences, poses a significant challenge applying to two-dimensional image data devoid of causal relationships. To address this issue, we propose the multiple scanning path mechanism, \textit{i.e.}, forward, reverse, and random shuffling.
To fuse the information flow from these diverse paths, the most straightforward method would be averaging. However, our objective is to adaptively activate the information derived from each path. Consequently, we have designed a gate to regulate the information flow from the various paths.
Tab. \ref{tab:ablation-path} illustrates the performance enhancements achieved through these designs. An increase in the number of paths correlates with an improvement in classification effectiveness. The gating mechanism also offers certain advantages over feature averaging.
It is important to note that we utilized average pooling features for classification in this instance. If we were to adopt a ViT-like class token design, the absence of a multi-path scheme would lead to a substantial decline in performance.

\begin{table}[!t] 
\centering
\caption{Effect of different scanning paths on performance.
}
\label{tab:ablation-path}
\vspace{-6pt}
\resizebox{\linewidth}{!}{
\begin{tabular}{cccc | *3{c}}
\toprule
Forward & Reverse & Shuffle & Mean/Gate  & P & R & F1
\\
\midrule
\checkmark& & &  - & 88.14 & 87.11 & 87.24
\\
\checkmark & \checkmark & & Mean & 89.55 & 88.45 & 88.61 
\\
\checkmark &\checkmark & & Gate & 90.83 & 89.87 & 90.07
\\
\checkmark & \checkmark & \checkmark & Gate & 92.02 & 91.53 & 91.66
\\
\bottomrule
\end{tabular}
}
\end{table}

\subsubsection{Effect of Positional Encoding}

To enhance RSMamba with the capacity to model relative spatial relationships, we incorporate position encoding into the flattened image sequence. Tab. \ref{tab:ablation-pe-token} delineates the influence of the presence, absence, and type of position encoding on the classification performance. The lack of position encoding leads to a degradation in performance, whereas both Fourier encoding and learnable encoding contribute to performance enhancements. It should be noted that, given RSMamba's ability to restore the tokens of different paths to their original order, the impact of the presence or absence of position encoding is somewhat mitigated. However, the integration of position encoding can still yield a slight incremental improvement.

\subsubsection{Effect of the Number of Tokens} 

RSMamba's proficient capability in global feature abstraction significantly alleviates the complications associated with the length of tokens. As a result, in this paper, we employ an overlapping image patch division method. Tab. \ref{tab:ablation-pe-token} elucidates the effects of the presence or absence of overlap, as well as the enlargement of image size. The division of image patches with overlap allows each token to encapsulate more exhaustive information, thereby leading to an enhancement in performance. Augmenting the image size facilitates the inclusion of more details, which correspondingly yields substantial performance gains. The linear modeling complexity employed by SSM enables a considerable increase in sequence length, even under conditions constrained by resources.

\begin{table}[!t] 
\centering
\caption{Effect of positional encoding.
}
\label{tab:ablation-pe-token}
\vspace{-6pt}
\resizebox{\linewidth}{!}{
\begin{tabular}{c| c | *3{c}}
\toprule
\multicolumn{2}{c|}{Design}  & P & R & F1
\\
\midrule
\multirow{3}{*}{PE} & None & 90.64 & 90.22 & 90.25
\\
& Fourier & 91.62 & 90.85 & 91.04
\\
& Learnable & 92.02 & 91.53 & 91.66
\\
\midrule
\multirow{3}{*}{Token} & $224^2\text{px}, k=16, s=16$ & 89.93 & 89.31 & 89.38
\\
& $224^2\text{px}, k=16, s=8$& 92.02 & 91.53 & 91.66
\\
& $384^2\text{px}, k=16, s=8$& 92.75 & 91.98 & 92.16
\\
\bottomrule
\end{tabular}
}
\end{table}

\section{Discussion and Conclusion}

In this paper, we introduce a novel state space model for remote sensing image classification, referred to as RSMamba. RSMamba concurrently harnesses the advantages of CNNs and Transformers, specifically their linear complexity and global receptive field. We introduce a dynamic multi-path activation mechanism to alleviate the limitations of unidirectional modeling and position insensitivity inherent in the vanilla Mamba. RSMamba maintains the internal structure of the Mamba and offers the flexibility to easily expand parameters to accommodate various application scenarios. Experimental evaluations conducted on three distinct remote sensing image classification datasets demonstrate that RSMamba can outperform other state-of-the-art classification methods based on CNN and Transformer. Consequently, RSMamba exhibits considerable potential to serve as the backbone network for next-generation visual foundation models.

\bibliographystyle{IEEEtran}
\bibliography{IEEEabrv,myreferences}
% \bibliography{IEEEabrv,../bib/paper}
%
% \section{Simple References}
% You can manually copy in the resultant .bbl file and set second argument of $\backslash${\tt{begin}} to the number of references
%  (used to reserve space for the reference number labels box).

% \begin{thebibliography}{1}

% % \bibitem{ref1}
% % {\it{Mathematics Into Type}}. American Mathematical Society. [Online]. Available: https://www.ams.org/arc/styleguide/mit-2.pdf

% \end{thebibliography}

\end{document}